# CAFCT-NET: A CNN-TRANSFORMER HYBRID NETWORK WITH CONTEXTUAL AND ATTENTIONAL FEATURE FUSION FOR LIVER TUMOR SEGMENTATION

*Ming Kang, Chee-Ming Ting, Fung Fung Ting, Raphaël C.-W. Phan*

School of Information Technology, Monash University, Malaysia Campus

## ABSTRACT

Medical image semantic segmentation techniques can help identify tumors automatically from computed tomography (CT) scans. In this paper, we propose a Contextual and Attentional feature Fusions enhanced Convolutional Neural Network (CNN) and Transformer hybrid network (CAFCT-Net) for liver tumor segmentation. We incorporate three novel modules in the CAFCT-Net architecture: Attentional Feature Fusion (AFF), Atrous Spatial Pyramid Pooling (ASPP) of DeepLabv3, and Attention Gates (AGs) to improve contextual information related to tumor boundaries for accurate segmentation. Experimental results show that the proposed model achieves a mean Intersection over Union (IoU) of 76.54% and Dice coefficient of 84.29%, respectively, on the Liver Tumor Segmentation Benchmark (LiTS) dataset, outperforming pure CNN or Transformer methods, e.g., Attention U-Net and PVTFormer.

## 1. INTRODUCTION

Medical image semantic segmentation is an important branch of computer vision tasks that can support clinical diagnosis. Convolutional neural networks (CNNs), typically U-Net [1] series, have greatly promoted the development of various semantic segmentation, such as H-DenseUNet [2], R2U-Net [3], DeepLabv2 [4], DeepLabv3 [5], Deeplabv3+ [6], Attention U-Net [7], and so on. U-Net, which has been widely used in medical image semantic segmentation, is based on an encoder-decoder structure and uses skip connections to reduce the fusion of different features. As its successor, UNet++ [8, 9] redesigned skip connections to combine features of different semantic scales within the decoder sub-networks, resulting in a highly adaptable feature fusion method. DeepLabv2, Deeplabv3, and Deeplabv3+ employ Atrous Spatial Pyramid Pooling (ASPP) to capture contextual information at multiple scales. In order to highlight salient features that are passed through the skip connections, Attention Gates (AGs) are used in the Attention U-Net. H-DenseUNet explores intra-slice and inter-slice features for liver and tumor segmentation and achieves competitive performance on liver tumor segmentation.

Transformer [10] was originally proposed for natural language processing as a model architecture containing self-attention in an encoder-decoder architecture. The roles of CNN and Transformer in feature extraction are different and complementary. The weight parameters of CNN trained in the kernel can extract the characteristics of the elements in the mixed response field. In contrast, Transformer adaptively extracts and blends features between all patches by obtaining the similarity of all patch pairs through the dot product between patch vectors. This gives the Transformer an effective global receptive field and reduces the inductive bias of the model. Therefore, Transformer has more powerful generalization capabilities than CNN and multi-layer perceptron-like structures. However, the low sensing bias and strong global receptive field make it difficult for converter models to adequately capture the critical local details of a specific task. In addition, as the Transformer model continues to deepen, global features continue to mix and converge, leading to distraction. These make it difficult for the Transformer model to accurately predict the information in the dense prediction task of detailed semantic segmentation in the dense prediction task. CTC-Net [11] is a CNN and Transformer complementary network that fuses mutual and cross-domain information in the feature extraction by two different encoders and achieves superior performance on multi-organ and cardiac segmentation. PVTFormer [12] that utilizes Pyramid Vision Transformer (PVT) v2 [13] as a backbone generates high-quality liver tumor segmentation masks by enhancing semantic features with pure Transformer encoder and pure CNN decoder architecture, which is considered as a pure Transformer-based segmentation model in the CNN-Transformer fusion's setting of this work.

In this paper, we propose a novel segmentation network called CAFCT-Net, which combines CNN and Transformer to leverage contextual attention for learning not only local and global information but also contextual knowledge. First, we use two branch backbone networks, CNN and Transformer, to extract features. In order to better fuse the two different types of extracted features, we design a novel Attentional Feature Fusion (AFF) module incorporating attention mechanisms, which is used to collect global information, capture channel-wise relationships, and improve representation ability. To avoid the different contexts between lesions in different periods, we introduce ASPP for multiscale lesions to enrich semantic information and detailed information related to lesion boundaries. One common application of attention mechanisms is in image segmentation tasks, where the model needs to identify and outline specific objects or regions within an image. We utilize AGs to help the model allocate more attention to important features while suppressing irrelevant or distracting information. In addition, we adopt a Binary Cross-Entropy (BCE) Dice loss. The main contributions of this paper are summarized as follows:

1) The backbone network of the proposed model incorporates the local and global features of CNN and Transformer through an attentional feature fusion module so that the model can learn context information, thereby comprehensively improving the performance of the liver tumor segmentation.

2) We design a novel attentional feature fusion module to comprehensively integrate the extracted local and global features, which is different from simple add and concatenation operations. The AFF module adaptively fuses semantic feature information based on the learned weighted contextual information of the two features.

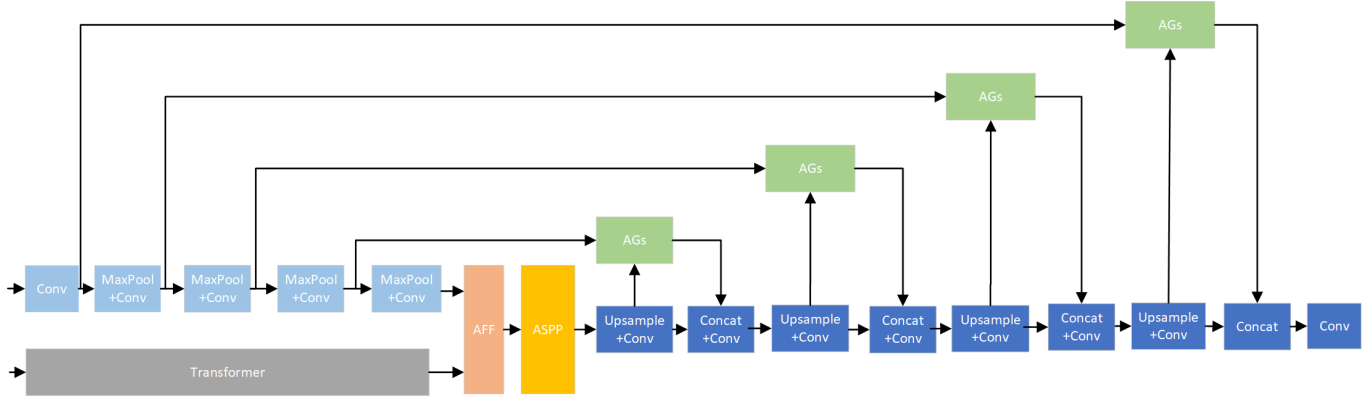

**Fig. 1**. Overview of CAFCT-Net. Conv represents the convolutional neural layer. AFF represents the Attentional Feature Fusion module. ASPP represents the Atrous Spatial Pyramid Pooling module. AGs represent the Attention Gates module. Concat represents concatenation.

3) We develop an innovative attention gate module functioning as skip connections to facilitate context learning, automatically adapting to target structures of diverse shapes and sizes. The proposed model trained with these attention gates inherently acquires the ability to diminish the significance of irrelevant areas in an input image while accentuating significant features crucial for semantic segmentation.

## 2. METHODS

In this section, we describe each module of the architecture of the proposed Contextual Attention Fusions enhanced Convolutional Neural Network (CNN) and Transformer hybrid network (CAFCT-Net) model as shown in Fig. 1. The CAFCT-Net has a two-branch encoder structure and attention-gated skip connections. The encoder is composed of CNN and Transformer branches, which are mainly responsible for feature extraction and downsampling. Then, contextual feature fusion is generated by three modules, i.e., AFF, ASPP, and AGs, in which the resolution of the output feature map of each layer is reduced by half compared to the previous layer, and the number of channels is doubled while the channel on the last layer remains unchanged. Finally, the convolutional decoder is designed for mapping the low-resolution feature map obtained by the encoder to the pixel-level prediction.

### 2.1. Combined CNN-Transformer Encoder

In the encoder part, we use two branches of encoders, namely a CNN-based encoder and a Transformer-based encoder. Although CNN has achieved unparalleled performance in numerous medical image segmentation tasks, it lacks efficiency in capturing global contextual information. Existing works attempt to capture global information by generating a very large receptive field, which requires continuous downsampling and stacking of convolutional layers until it is deep enough. This brings several disadvantages: First, the training of very deep networks will be affected by the problem of diminishing feature reuse, in which low-level features are washed out by successive multiplication operations. Secondly, as the spatial resolution gradually decreases, it suffers the loss of local information crucial for dense prediction tasks such as pixel-by-pixel segmentation. Transformers, used for sequence-to-sequence prediction, have emerged as alternative architectures with inherent global self-attention mechanisms but may have limited localization capabilities due to insufficient underlying details. Therefore, CNN and Transformer dual encoders are better solutions to improve the efficiency of modeling global context while maintaining robust capture of low-level details.

To leverage the advantages of CNNs plus Transformers in medical image semantic segmentation, CAFCT-Net combines convolution with transformers to capture local and global information simultaneously. CNN's branch progressively expands the receptive field, capturing features on a local and global scale. Meanwhile, the Transformer branch initiates with global self-attention, ultimately restoring local details. These two branches contribute to the extraction of features. In the CNN branch, features are progressively downsampled, and hundreds of layers are employed to obtain the global context of features. In the Transformer branch, the input image is first evenly divided into patches, and then the patches are flattened and passed into a linear embedding layer with output dimension.

### 2.2. Attentional Feature Fusion

To integrate multi-level features respectively extracted by CNN-based and Transformer-based encoders to segment liver tumors, we design an attentional feature fusion module to mine contextual semantics for improved lesion segmentation. Dai et al. [14] proposed the attentional feature fusion framework that aggregates local and global contextual features given at different scales. Inspired by this work, we build an attentional feature fusion module leveraging Squeeze-and-Excitation Networks (SENet) [15] structure to gather multiscale feature contexts to address issues arising from variations in scale.

As shown in Fig. 2, we first concatenate features $F_{CNN}$ and $F_{Trans}$ from CNN and Transformer branches respectively. and then interactively input the output to SENet blocks to explicitly model the interdependencies between the channels of its convolutional feature. The SENet block is divided into two operations that are associated with attention and gating mechanisms. First, the squeeze operation, which represents the global distribution of the response on the feature channel and enables the layers close to the input to obtain the global receptive field, employs a global pooling operation to compress the 3D of $H \times W \times C$ to 1D vector $1 \times 1 \times 1$. Compared with the ordinary convolution capturing the size of the local receptive field at the lower level, the receptive field becomes wide avoiding errors via squeeze operation. The squeeze operation performs feature compression along the spatial dimension and turns each two-dimensional

2971

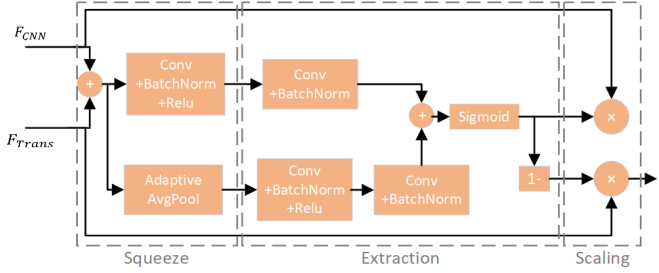

**Fig. 2**. The structure of attentional feature fusion module. It can be divided into squeeze, extraction, and scaling parts according to the SENet block.

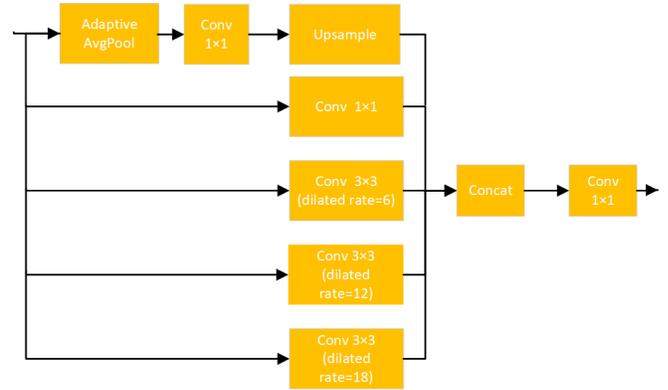

**Fig. 3**. The structure of atrous spatial pyramid pooling module. It employs atrous convolutions with three different dilated rates that extract the different scales of feature information.

feature channel into a real number that has a global receptive field to some extent. The output dimension matches the number of input feature channels so that the information on the global receptive field can be obtained. Second, the excitation operation, which is similar to the gating function, gives each part a correlation weight value. The importance of each feature channel is automatically obtained through excitation operation. Thus, useful features are improved, and features that are not very useful for the current task are suppressed simultaneously according to their importance.

### 2.3. Atrous Spatial Pyramid Pooling

The boundary information of image segmentation has a great influence on the final segmentation results, which may cause the inaccurate segmentation of the object boundaries, leading to pixel-by-pixel misclassification. We designed a simple yet effective edge-aware module. The atrous spatial pyramid pooling module, first proposed in DeepLabv2, integrates low-level local boundary information and high-level global position information to explore edge semantics related to object boundaries under explicit boundary supervision [16]. We introduce the ASPP module to combine boundary features with liver tumor features at each layer to guide representation learning for image segmentation and make the model pay more attention to the structure and details of the tumors. Thus, multi-level fusion features of the former module AFF are gradually aggregated from top to bottom to predict the boundary information of segmented liver tumors.

On the other hand, due to the different parameters and techniques set by the Computed Tomography (CT) equipment during the image acquisition process, and according to the label data of the liver tumor CT image, the lesions on livers have different sizes and shapes, which results in the network model not being able to effectively extract feature information, and hence sub-optimal segmentation performance. The ASPP module uses atrous convolution and global average pooling with different dilated rates to extract feature information at different scales in CT images, which can effectively improve segmentation performance.

Fig. 3 illustrates the structure of the ASPP module. The first step of the ASPP module requires five convolutional layers to operate in parallel: the first convolution layer reduces the size of the original image in order to obtain the global context information via adaptive average pooling performed on each channel of the input feature map, and then a new feature map is formed through $1 \times 1$ convolution; the features are processed using upsample to restore the size after convolution is also performed. The second convolutional layer uses a $1 \times 1$ convolution kernel to perform a convolution operation on a mixture of features. The third to fifth convolutional layers allow the input feature map to pass through atrous convolutions with dilated rates of 6, 12, and 18 and a convolution kernel size of $3 \times 3$, respectively. The atrous convolution can increase the field of view using different dilated rates. When the kernel size is small, the receptive field is expanded to extract more feature information. Then, output feature maps from each parallel convolution layer are spliced to increase the correlation between feature maps of different scales because the original feature maps are spliced in the channel dimension. So the obtained channel dimension is the original five times the feature map; then use $1 \times 1$ convolution to reduce the dimensionality of the obtained feature map. Finally, a batch normalization operation is applied to produce the final feature map of the ASPP module.

### 2.4. Attention Gates

Attention gates, firstly proposed in [7, 17, 18], automatically learn to focus on different shapes and the size of the object structure as shown in Fig. 4. Attention mechanisms are employed to spotlight or accentuate specific portions of the input data, enabling the model to concentrate on pertinent information. The AGs can be integrated into architectures to improve the model's performance by selectively attending to diverse regions within the input. Since semantic contextual differences among various-scale feature maps can impact feature fusion in skip connections, we introduce skip connections to the proposed model with AGs modules. These modules aid in the fusion of features between semantically misaligned feature maps, leading to a notable enhancement in segmentation results.

AGs exhibit distinctive characteristics with features and gated signals serving as inputs. This involves employing $1 \times 1$ convolution to standardize their sizes, followed by an element-wise addition to obtain superimposed features. Subsequently, activation is applied, and a grid of attention coefficients is obtained using bilinear interpolation. In the process of identifying significant image regions and fine-tuning feature responses, the input features are multiplied element-wise with the attention coefficient derived from the gating signal obtained at a coarser scale—this activation and contextual information aid in selecting spatial regions. The attention gate is operational during forward and backward propagation, implementing neuron activation. In the reverse process, the gradient from the background area is weighted downward. The model inherently learns to suppress irrelevant image areas while accentuating task-relevant salient features. These features can be integrated into standard image analysis models based on convolutional networks, enhancing sensitivity, prediction accuracy, and suitability for dense label prediction.

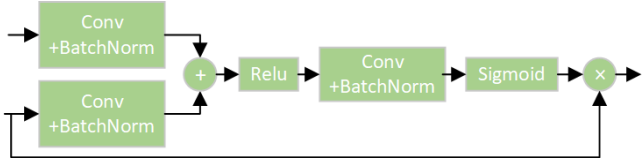

**Fig. 4**. The structure of attention gates module. AGs filter the features propagated through the skip connections as shown in Fig. 1.

As shown in Fig. 1, the CNN layers divide features evenly into four feature maps and integrate features from adjacent layers via AGs in the decoder part. The proposed model leverages high-level abstract features obtained from the ASPP module as gating signals. These signals, along with the input features, are fed into the Attention Gate, where attention coefficients are computed based on activation and context information derived from the gating signal. Grid resampling is executed through bilinear interpolation. Utilizing attention coefficients, the input features are updated to pinpoint spatial regions pertinent to the designated task. This process filters and extracts crucial spatial features, which are then merged with refined features obtained by quadruple bilinear upsampling of the gated signal. The final output is ultimately generated through a combination of convolutional layers and bilinear upsampling.

### 2.5. Multi-Loss Function

For achieving precise semantic segmentation, we employ a multi-loss approach known as BCE-Dice loss, which integrates binary cross-entropy loss and Dice loss. Dice loss provides a global overview, while BCE conducts a detailed examination pixel by pixel, offering complementary perspectives. This combination introduces variability in the loss, maintaining stability attributed to BCE. However, it's important to note that BCE-Dice loss may not be suitable for segmenting small-scale objects, such as a $10 \times 10$ object in a $512 \times 512$ image, as the BCE loss is impacted in such scenarios, unlike Dice loss.

## 3. EXPERIMENTS

### 3.1. Dataset

We evaluated the proposed CAFCT-Net on the Liver Tumor Segmentation Benchmark (LiTS) [19] dataset. The original LiTS dataset comprises 201 abdominal CT training scans with contrast enhancement, sourced from seven distinct clinical institutions where trained radiologists manually segmented the livers and lesions in the provided training CT scans. The training data set contains 131 scans, and the test set has 70 scans.

However, we used an open-access 2D slice dataset [20] with a total of 58,638 images converted from 131 CT 3D scans because the 70 scans in the test data set of the original LiTS dataset are not annotated. Tumor lesions and liver organs are binarily masked (i.e., white mask and black background) and saved as images, resulting in 58,638 lesion-mask/tumor-mask and 58,638 liver-mask images, respectively. Unlike 3D scans, some 2D slices contain both lesion and liver masks or either lesion or liver mask, and others contain no masks. The size of each 2D image is resized to $256 \times 256$ pixels. The train and test sets have been divided by the author of the 2D slice dataset and are only parts of all slice/mask images. Table 1 shows the number of samples/instances in the train and test sets.

**Table 1**. Number of samples/instances in the train and test sets.

| Set | 3D Scan | 2D Slice | Lesion Mask | Liver Mask |
|---|---|---|---|---|
| Train | 79 | 38,523 | 34,013 | 26,394 |
| Test | 11 | 3,038 | 2,675 | 1,838 |
| All | 90 | 41,561 | 36,688 | 28,232 |

**Table 2**. Performance comparison of CAFCT-Net and other methods on the LiTS dataset for lesion segmentation. The unit of mIoU and Dice coefficient are the percentage (%).

| Methods | mIoU | Dice |
|---|---|---|
| H-DenseUNet [2] | 68.30 | 78.64 |
| DeeplabV3+ [6] | 73.98 | 82.05 |
| Attention U-Net [7] | 65.92 | 76.10 |
| PVTFormer [12] | 73.13 | 82.87 |
| **CAFCT-Net (Ours)** | **76.54** | **84.29** |

### 3.2. Implementation Details

Our CAFCT-Net is based on the Pytorch deep learning framework. For the experimental setup, we tuned a number of hyperparameters, such as learning rate, batch size, weight decay rate, resize, etc. Ubuntu 20.04 was utilized on both the training and testing computers, and single NVIDIA® RTX® 4090 graphics cards, each with 24GB of video memory, were used to set up the hardware. For training, we employed the small batch stochastic gradient descent (SGD) method with a batch size of 8 and a learning rate of 0.001. Adam optimization and SGD techniques were compared, and it was found that SGD typically delivers greater performance, although Adam converges more quickly.

### 3.3. Results

The comparison of the proposed CAFCT-Net with previous advanced deep learning methods on the LiTS dataset with 2D slices for lesion segmentation (i.e., liver tumor segmentation) is presented in Table 2. The mIoU and Dice coefficient of our proposed CAFCT-Net are 76.54% and 84.29%, respectively. Compared with Attention U-Net, ours achieved mIoU improvement of absolute 10.62% and 8.19% on the Dice coefficient evaluation metric. Experiment results show that our proposed methods can achieve better semantic segmentation performance than the typical use of CNN or Transformer method alone on liver tumor images.

To validate the generalizability of our proposed method, we conducted experiments on the same dataset for liver segmentation. Table 3 shows that our proposed CAFCT-Net achieves 91.92% and 95.64% in mIoU and Dice coefficient, respectively, and surpasses the performance of U-Net, DeeplabV3+, Attention U-Net, and UNet++.

### 3.4. Ablation Study

In the ablation study, we conducted experiments by removing only one module once from the baseline model to investigate the performance of each module. Table 4 displays our ablation study results towards different modules in CAFCT-Net in the above lesion segmentation tasks. Results show that the performance of the original CAFCT-Net decreases without whatever module. When the Transformer branch is replaced by a CNN branch same as the original one,

2973

**Table 3**. Performance comparison of CAFCT-Net and other methods on the LiTS dataset for liver segmentation. The units of mIoU and Dice coefficient are the percentage (%).

| Methods | mIoU | Dice |
|---|---|---|
| U-Net [1] | 89.31 | 93.60 |
| DeeplabV3+ [6] | 91.77 | 95.43 |
| Attention U-Net [7] | 89.65 | 94.09 |
| UNet++ [8, 9] | 90.05 | 94.43 |
| **CAFCT-Net (Ours)** | **91.92** | **95.64** |

**Table 4**. Ablation study of CAFCT-Net. The bracketed values represent the amount of decrease or increase compared with the original CAFCT-Net. The units of mIoU and Dice coefficient are the percentage (%). w/o denotes without.

| Methods | mIoU | Dice |
|---|---|---|
| w/o Transformer | 74.12 (-2.42) | 82.74 (-2.64) |
| w/o AFF | 76.15 (-0.39) | 84.51 (0.22) |
| w/o ASPP | 74.95 (-1.59) | 83.63 (-0.66) |
| w/o AGs | 74.03 (-2.51) | 83.41 (-0.12) |

the mIoU and Dice coefficient are both dramatically reduced. This demonstrates that the features extracted by the Transformer branch play a vital role in our proposed CAFCT-Net. Meanwhile, the performance occurs a big drop when the AGs are removed, which AGs in CAFCT-Net work well to guide the model's attention to important regions and substantially enhance the representational power of the model. The ASPP module has a better effect on fusing features from both CNN and Transformer branches than the AFF module because the performance declines much larger when ASPP is removed.

## 4. CONCLUSION

We developed a novel contextual and attentional fusion network, CAFCT-Net, for liver tumor segmentation by efficiently combining features extracted from CNN and Transformer's branches. The attentional feature fusion and attention gate modules are used to suppress irrelevant contextual information in the encoders and increase the weight of the liver tumor channel. Combined with the atrous spatial pyramid pooling module, our model improves the segmentation accuracy of tumor boundaries in different sizes. Experimental results on the LiTS dataset show that our method achieved better performance on liver tumor semantic segmentation compared to the use of CNN or Transformer method alone in previous studies.